\documentclass[11pt,a4paper]{article}
\usepackage[hyperref]{acl2018}
\usepackage{cleveref}
\usepackage{times}
\usepackage{latexsym}

\usepackage{amsmath}
\usepackage{amssymb}
\usepackage{color}
\usepackage{caption}
\usepackage{import}
\usepackage{enumerate}
\usepackage{enumitem}
\usepackage{mathrsfs}
\usepackage{graphicx}
\usepackage{multirow}
\usepackage{CJKutf8}
\usepackage{subcaption}
\usepackage{wrapfig}

\usepackage{url}
\aclfinalcopy % Uncomment this line for the final submission
\title{How Do Source-side Monolingual Word Embeddings\\ Impact Neural Machine Translation?}

\author{Shuoyang Ding$\dagger$\quad\quad Kevin Duh$\dagger$$\ddagger$\\
  $\dagger$ Center for Language and Speech Processing \\
  $\ddagger$ Human Language Technology Center of Excellence\\
  Johns Hopkins University \\
  Baltimore, Maryland, United States \\
  {\tt dings@jhu.edu\quad kevinduh@cs.jhu.edu} \\}

\date{}

\definecolor{pms651c}{RGB}{167,188,214}
\definecolor{pms584c}{RGB}{210,215,85}
\definecolor{pms564c}{RGB}{134,200,188}
\definecolor{pms1215c}{RGB}{251,216,114}
\begin{document}
\maketitle
\maketitle

\begin{abstract}
  Using pre-trained word embeddings as input layer is a common practice in many natural language processing (NLP) tasks, but it is largely neglected for neural machine translation (NMT). In this paper, we conducted a systematic analysis on the effect of using pre-trained source-side monolingual word embedding in NMT. 
 % Our experiments indicate that pre-trained source-side monolingual word embedding cannot directly benefit NMT and some form of adaptation on bilingual data needs to be done to make it helpful.
We compared several strategies, such as fixing or updating the embeddings during NMT training on varying amounts of data, and we also proposed a novel strategy called {\it dual-embedding} that blends the fixing and updating strategies. 
%We concluded that adaptation on bilingual data is necessary for these embeddings to be helpful. Several different adaptation strategies, including our own novel adaptation strategy called dual-embedding, is compared. 
%Overall, we observed most empirical improvements when incorporating word embedding that's been trained on monolingual data that is of larger scale compared to the bilingual training data, and we observed complementary effect between the pre-trained word embedding and the embedding learned from scratch.
Our results suggest that pre-trained embeddings can be helpful if properly incorporated into NMT, especially when parallel data is limited or additional in-domain monolingual data is readily available.
%adaptation on bilingual data is necessary for these pre-trained embeddings to be helpful, and they should be used with NMT whenever parallel data is limited or additional in-domain monolingual data is readily available.
\end{abstract}

% \import{\sectiondir}{todo.tex}
\section{Introduction}

% Neural machine translation (NMT) \cite{Bahdanau:2014vz} \cite{Anonymous:WiX-kbXc} systems achieved high performances in several different evaluations \cite{Bojar:2016ug}\cite{Bojar:2017td}. However, NMT systems have no stand-alone language model module and hence is not able to directly make use of monolingual data. As \cite{Och:2004hu} indicated that monolingual data helps with translation quality, it's a natural hypothesis that improvements could be achieved if monolingual data could be successfully incorporated into NMT systems.

Leveraging the information encoded in pre-trained monolingual word embeddings \cite{bengio2003neural,mikolov2013distributed,pennington2014glove,bojanowski2016enriching} is a common practice in various natural language processing tasks, for example: parsing \cite{Dyer:2015tt} \cite{Kiperwasser:2016uza}, relation extraction \cite{peng2017cross}, natural language understanding \cite{cheng2016long}, sequence labeling \cite{ma2016end}, and so on. While research in neural machine translation (NMT) seeks to make use of monolingual data by techniques such as back-translation \cite{Sennrich:2015tp}, the use of pre-trained monolingual word embeddings does not seem to be the standard practice. 
%only a very limited amount of work exploits pre-trained monolingual word embeddings.

In this paper, we study the interaction between the \textbf{source-side} monolingual word embeddings with NMT. Our goal is to understand whether they help translation accuracy, and how to integrate them into the model in the best way. Specifically, we seek to answer questions such as: (1) Should the embeddings be fixed or updated during NMT training? (2) Are embeddings more effective when bilingual training data is limited? (3) Does the amount or domain of monolingual data used to train the embeddings affect results significantly? We answered these questions with experiments on varying amounts of bilingual training data drawn from Chinese-English and German-English tasks. 

Additionally, we proposed a simple yet effective strategy called \textit{dual embedding}, which combines a fixed pre-trained embedding with a randomly initialized embedding that is updated during NMT training. We found that this strategy generally improves BLEU under various data scenarios, and is an useful way to exploit pre-trained monolingual embeddings in NMT. 

%The reason for focusing on the source-side is that most of the current works that seeks to make use of source-side monolingual data involves auto-encoder-style training, which is currently slow and sensitive to hyper-parameter choices. We, instead, prefers method that achieves the same goal while incurring minimum change over the original system.

% We concluded from preliminary experiments that using unadapted monolingual word embedding is not very helpful with neural machine translation. We hereby investigated several different adaptation schema, including one we proposed which is named as \texttt{dual-embedding} throughout the paper. 

% The structure of this paper is as follows: section \ref{sec:related} summarizes the related works; section \ref{sec:dual} introduces the dual embedding model; section \ref{sec:exp} describes our experimental setup, results and our analysis; section \ref{sec:conclusion} concludes the paper.

% \texttt{\color{red} One paragraph briefly talking about result here}

% \texttt{\color{red} One paragraph that gives the outline here}

\section{Related Work} \label{sec:related}
% \noindent \textbf{Related Work}\quad
Various works have explored the use of monolingual data in NMT. On using target side monolingual data, \cite{gulcehre2015using} incorporated a pre-trained RNN language model, while \cite{Sennrich:2015tp} trained an auxiliary model with reverse translation direction to construct pseudo parallel data. \cite{Currey:2017ui} copied target monolingual data to the source side and mix it with real parallel data to train a unified encoder-decoder model. It should be noted that although some of these techniques are very popular in practice, they are only capable of incorporating \textbf{target-side} monolingual data, and hence they are outside the immediate scope of this paper.

%In addition, various works seek to incorporate pre-trained source-side word embedding into NMT systems. \cite{Gonzales:2017tt} used sense embedding to improve word sense disambiguation abilities of the translation system, but yielded limited improvements over the baseline. \cite{abdou2017variable} used pre-trained word embedding to help training converge faster, but did not analyze the effect of pre-trained word embedding on translation qualities. \cite{di2017monolingual} studied the effect of using source monolingual embedding for NMT at about the same time, focusing mostly on low resource settings. While their gated sum approach is also interesting, they did not empirically motivate why it is beneficial to combine pre-trained word embedding with bilingual NMT word embedding trained from scratch, instead of simply initializing certain weights with pre-trained word embedding, a gap filled by our study.

On using source side monolingual data, \cite{Cheng:2016tg,Zhang:2016uk} proposed self-learning for NMT, which learns the translation by reconstruction. Also, several works have shown promising results in incorporating pre-trained source-side word embedding into NMT systems. \cite{Gonzales:2017tt} used sense embedding to improve word sense disambiguation abilities of the translation system. Most similar to our work is \cite{abdou2017variable}, which used pre-trained word embedding to help training converge faster, and \cite{di2017monolingual}, which studied the effect of source monolingual embedding for NMT, but focus on low resource settings; further, their gated sum approach to combining embeddings has similar motivation to our dual embedding strategy. The results and analysis in our work verifies most of the conclusions in their paper, while extending them by examining both low and high resource settings and providing additional answers to question of when and how are embeddings beneficial. 

% Related to the spirit of our work in another aspect, \cite{Hill:2014ug} conducted analysis on word embeddings trained by neural machine translation systems and showed that the bilingual word representation extracted from a neural machine translation system is able to surpass several different monolingual or bilingual word representation models on word similarity tasks. \cite{faruqui2014improving} \cite{artetxe2016learning} \cite{artetxe2017learning} built bilingual word embeddings using monolingual data as a starting point by assuming certain algebraic properties of the representation space. Our work is an extension of these works geared toward the application of neural machine translation.
\section{Strategies for Incorporating Pre-trained Embeddings}

\subsection{Initializing Embedding}

We focus on simple strategies to incorporate pre-trained source-side monolingual embeddings in the NMT encoder.
The goal is to require minimum modification of the standard NMT model.
The simplest way is to initialize the NMT model with pre-trained embedding values -- for each word in the source sentence of the training bitext, we look-up its pre-trained embedding and initialize the NMT encoder with it.

After initialization, there are two strategies during training.
We can \textbf{fix} the embeddings to their initialized pre-trained values throughout training; in other words, NMT training optimizes parameters in the context encoding layers (e.g. LSTM), the attention layers, etc., but do not back-propagate into the embedding parameters. Example works adopting such strategy include \cite{Dyer:2015tt} and \cite{Kiperwasser:2016uza}.

Alternatively, we can \textbf{update} the embeddings with all other NMT model parameters during training. Here, the pre-trained word embeddings only provide an initialization that is different from the baseline of random initialization. 
Updating the initialization allows embeddings to adjust to values that are suitable for the NMT overall objective, whereas fixing embeddings potentially allows generalization to test words that are not observed in the training data. Example works adopting such strategy include \cite{ma2016end} and \cite{peng2017cross}.

\subsection{Dual Embedding} 
\label{sec:dual}

\begin{figure}
\centering
\hspace{-0.5cm}
\includegraphics[scale=0.5]{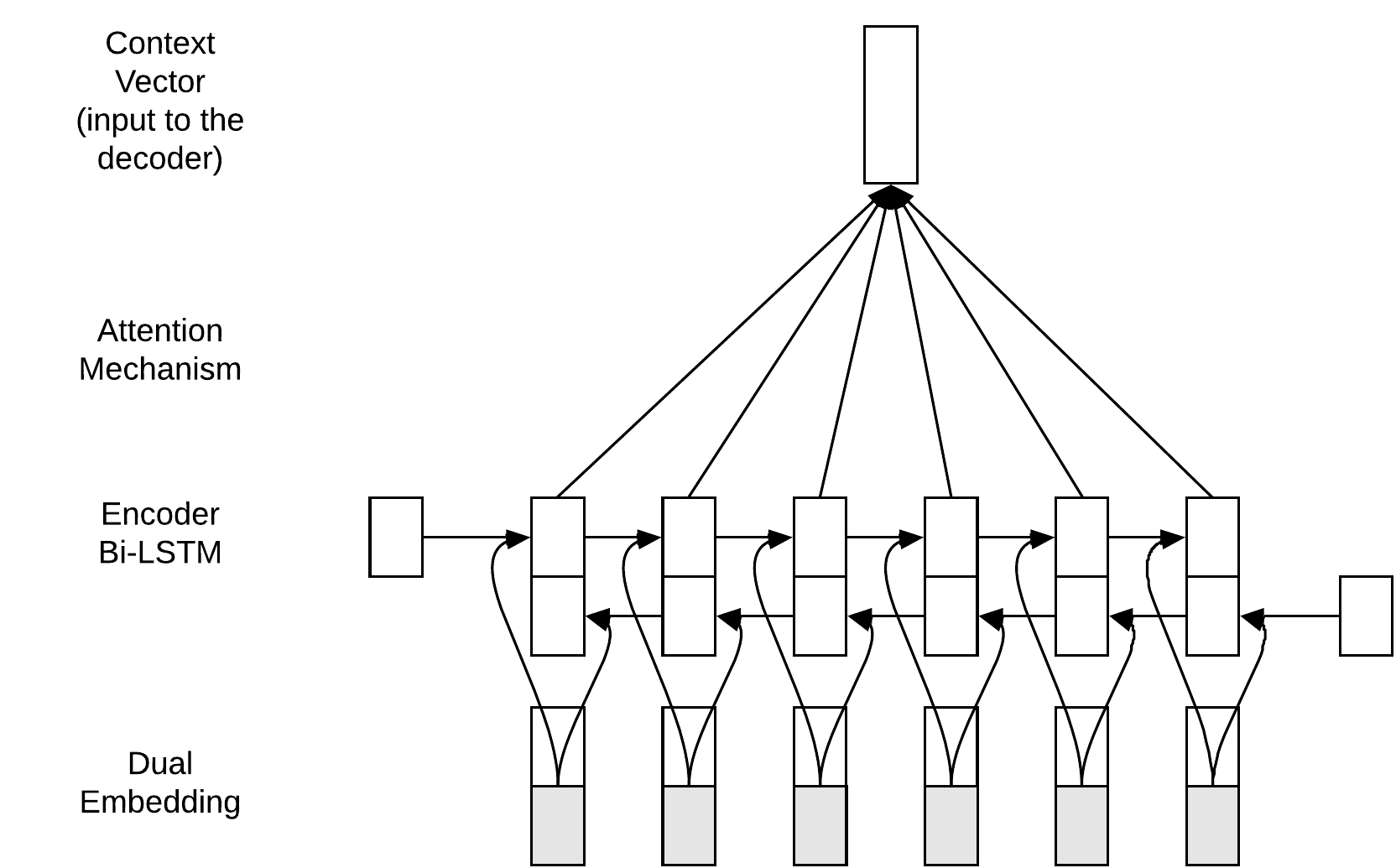}
\caption{Encoder part of NMT with dual embedding model. % The dashed lines and the shaded embeddings are the dual embedding model introduced in this work. Note that the shaded embeddings are fixed throughout the training process.
Note that the shaded embeddings are initialized with pre-trained embedding and fixed throughout the training process.
\label{fig:1}}
\end{figure}

To combine the benefit of both strategies introduced above, we propose {\it dual embedding}, as shown in Figure \ref{fig:1}. Basically, we augment the original encoder-decoder architecture of NMT model with an extra word embedding of same size, which is concatenated with the original word embedding. While the original word embedding is randomly initialized%\texttt{\color{red} what about initializing to zero?}
, this extra embedding will be initialized by a pre-trained monolingual word embedding and fixed through out training. The idea behind this architecture is that we would like to learn a correction term over the original monolingual word embedding, rather than rewriting the monolingual word embedding altogether. % \texttt{\color{red} Do we want to make this update minimal by changing the objective function to penalize too much update? For example, add a term $\mid\mid W_{fix} - W_{update}\mid\mid^2$.}
\section{Experiments} \label{sec:exp}

% \usepackage{multirow}
\begin{table}
% \centering
\hspace{-0.5cm}
\begin{tabular}{l|l|l|l|l}
\multirow{2}{*}{Sample Size} & \multicolumn{2}{l|}{ZH-EN} & \multicolumn{2}{l}{DE-EN} \\ \cline{2-5} 
                             & Pre-BPE      & BPE         & Pre-BPE     & BPE          \\ \hline
100,000                      & 15.1\%       & 1.73\%      & 24.0\%      & 2.41\%       \\ \hline
250,000                      & 10.4\%       & .809\%     & 17.5\%      & .894\%      \\ \hline
500,000                      & 8.49\%       & .667\%     & 14.4\%      & .480\%      \\ \hline
1,000,000                    & 6.21\%       & .398\%     & 11.6\%      & .332\%      \\ \hline
Unsampled                    & 5.11\%       & .284\%     & 7.36\%      & .0923\%    
\end{tabular}
\caption{Percentage of OOV Types in the Test Sets}
\label{tab:1}
\end{table}

% TODO: make the number & percentage separation an extra vertical bar
% \begin{table}
% \centering
% \begin{tabular}{l|l|l}
% \begin{tabular}[c]{@{}l@{}}Sentence\\ Pairs\end{tabular}   & ZH-EN        & DE-EN         \\ \hline
% 100,000   & 122 / 1.73\% & 261 / 2.41\%  \\ \hline
% 250,000   & 57 / 0.809\% & 97 / 0.894\%  \\ \hline
% 500,000   & 47 / 0.667\% & 52 / 0.480\%  \\ \hline
% 1,000,000 & 28 / 0.398\% & 36 / 0.332\%  \\ \hline
% All       & 20 / 0.284\% & 10 / 0.0923\% \\
% \end{tabular}
% \caption{Number of OOV Word Types and Percentage of OOV Tokens in the Test Sets}
% \label{tab:1}
% \end{table}

%\subsection{Setup}
We conducted experiments for Chinese-English (ZH-EN) and German-English (DE-EN) language pairs. For ZH-EN experiments, a collection of LDC Chinese-English data with about 2.01 million sentence pairs was used as training set, while NIST OpenMT 2005 and OpenMT 2008 dataset were used as development set and test set, respectively. 
For DE-EN experiments, we used WMT 2016 data for training, newstest2008 as development set, and newstest2015 as test set.
To investigate the effectiveness of pre-trained monolingual embedding on systems trained on different amount of bilingual data, we varied the amount of training data in the experiments by performing random sampling on the full parallel data; Table~\ref{tab:1} shows the number and percentage of OOV word types for each training subset.

The monolingual data used for ZH-EN experiments is the XMU LDC monolingual data provided in WMT 2017 news translation evaluation \cite{Bojar:2017td}, while for DE-EN the German news crawl 2016 dataset was used as the monolingual data.
To investigate whether monolingual data size is a significant factor in NMT translation quality,
we experimented with two different kinds of monolingual word embeddings:
\begin{itemize}[noitemsep,topsep=2pt]
\item \textbf{small}: only the source-side of the parallel corpus is used for pre-training
\item \textbf{extended}: additionally, the monolingual corpus is used for pre-training
\end{itemize}

BPE was applied \cite{Sennrich:2015um} for both parallel and monolingual corpus with operation number 49,500, while total vocabulary size was set to 50,000\footnote{To reduce variance introduced by BPE over different size of the training data, we used a single BPE model trained on the source-side monolingual data and a comparable sample of English news crawl 2016 monolingual data.}. 

We compared the following strategies to incorporate pre-trained monolingual word embeddings.
\begin{itemize}[noitemsep,topsep=2pt]
\item \textbf{fixed} initialization: the source-side word embedding is fixed during training
\item \textbf{update} initialization: the source-side word embedding is updated during training
\item \textbf{dual} embedding: one half of the word embedding parameter contains the fixed pre-trained vector, while the other half is updated during training and acts as a correction term. 
\end{itemize}
These were compared to the \textbf{baseline} of using random initialization.

A modified fork of OpenNMT-py\footnote{\tt https://github.com/shuoyangd/OpenNMT\-py/tree/mono\_emb} \cite{klein2017opennmt} 
was used to run all NMT experiments, while fastText \cite{bojanowski2016enriching} was used to build  all the pre-trained monolingual word embedding, with embedding dimension 300. We used 2-layer LSTM for both the encoder and decoder, and the hidden dimension of LSTM was set to 1024. We updated the parameter with Adadelta \cite{zeiler2012adadelta} with learning rate 1.0, $\epsilon = 1e-6$ and $\rho = 0.95$, and performed dropout on all the LSTM layers and embedding layers with a rate of 0.2. We trained all of our models for 20 epochs, the best model selected by the perplexity on the validation set will be used to decode on test set.

We evaluated our models with uncased BLEU calculated using \texttt{multi-bleu.perl} that comes with the Moses decoder \cite{Koehn:2007vu}. We also performed pairwise significance testing \cite{koehn2004statistical} between some key experimental setups.

% plain initialization and fixed initialization. In both alternative adaptation strategies, the source side word embedding of the NMT model is initialized with the pre-trained monolingual word embedding. This word embedding is then updated by the parallel data under plain initialization and remain fixed under fixed initialization. We also trained two different kind of monolingual word embedding for initialization: the {\tt small} word embedding is trained only on the source side of the parallel corpus, while the {\tt full} word embedding is also trained on the monolingual data.

% In the domain adaptation experiments, as the all-domain monolingual data is large, we experimented with monolingual data sampling methods, including random sampling and Moore-Lewis sampling \cite{moore2010intelligent}. When we use Moore-Lewis sampling, we use the in-domain development set to train the in-domain 5-gram KenLM \cite{heafield2011kenlm} model and the all-domain monolingual data to train the out-of-domain 5-gram KenLM model. For both samspling methods, we sample 1 million monolingual sentences to build monolingual word embedding with fastText.

\begin{figure*}
\centering
\begin{subfigure}[b]{0.40\textwidth}
\centering
\includegraphics[width=\textwidth]{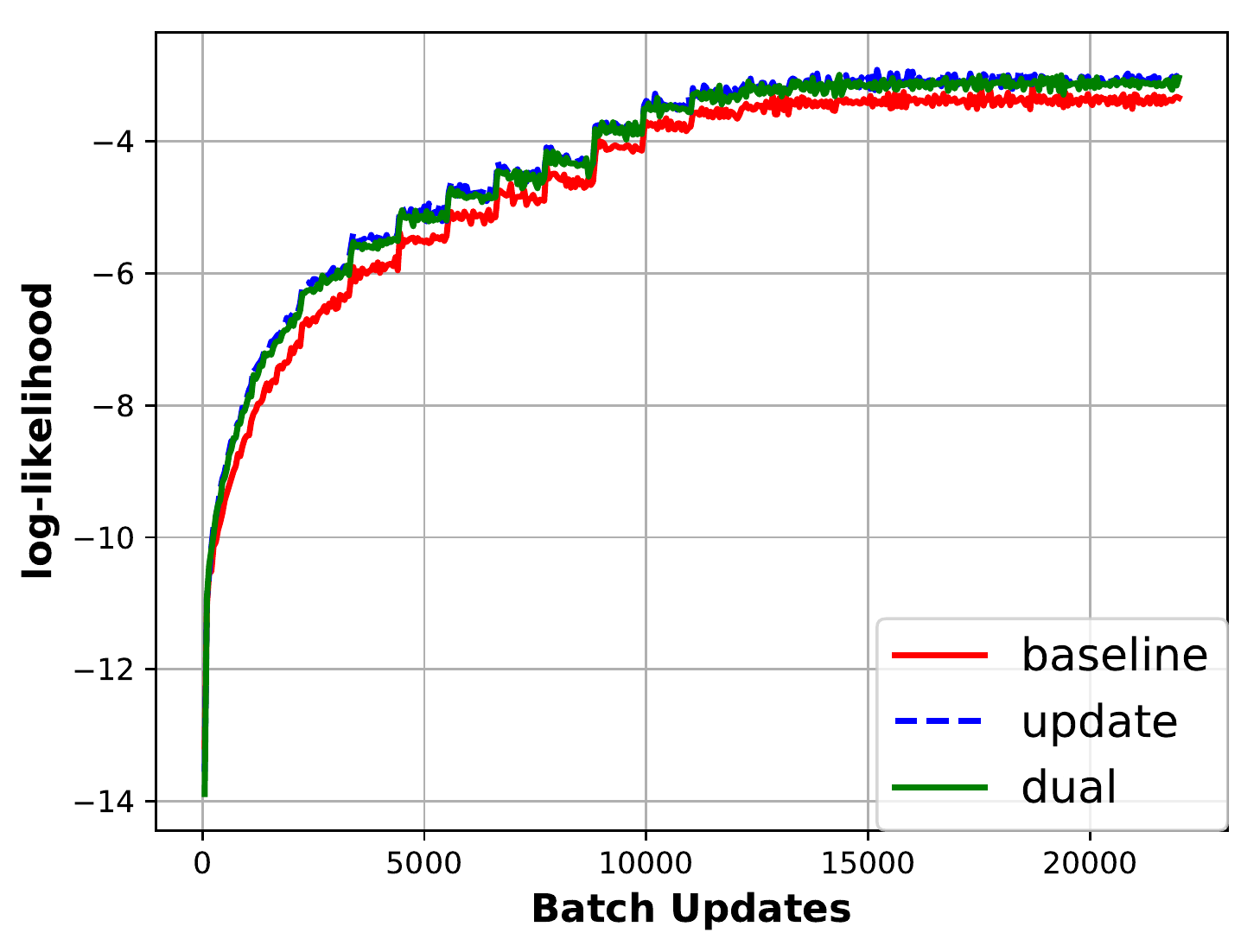}
\caption{Training Curve on 100k Sample}
\end{subfigure}
\begin{subfigure}[b]{0.40\textwidth}
\centering
\includegraphics[width=\textwidth]{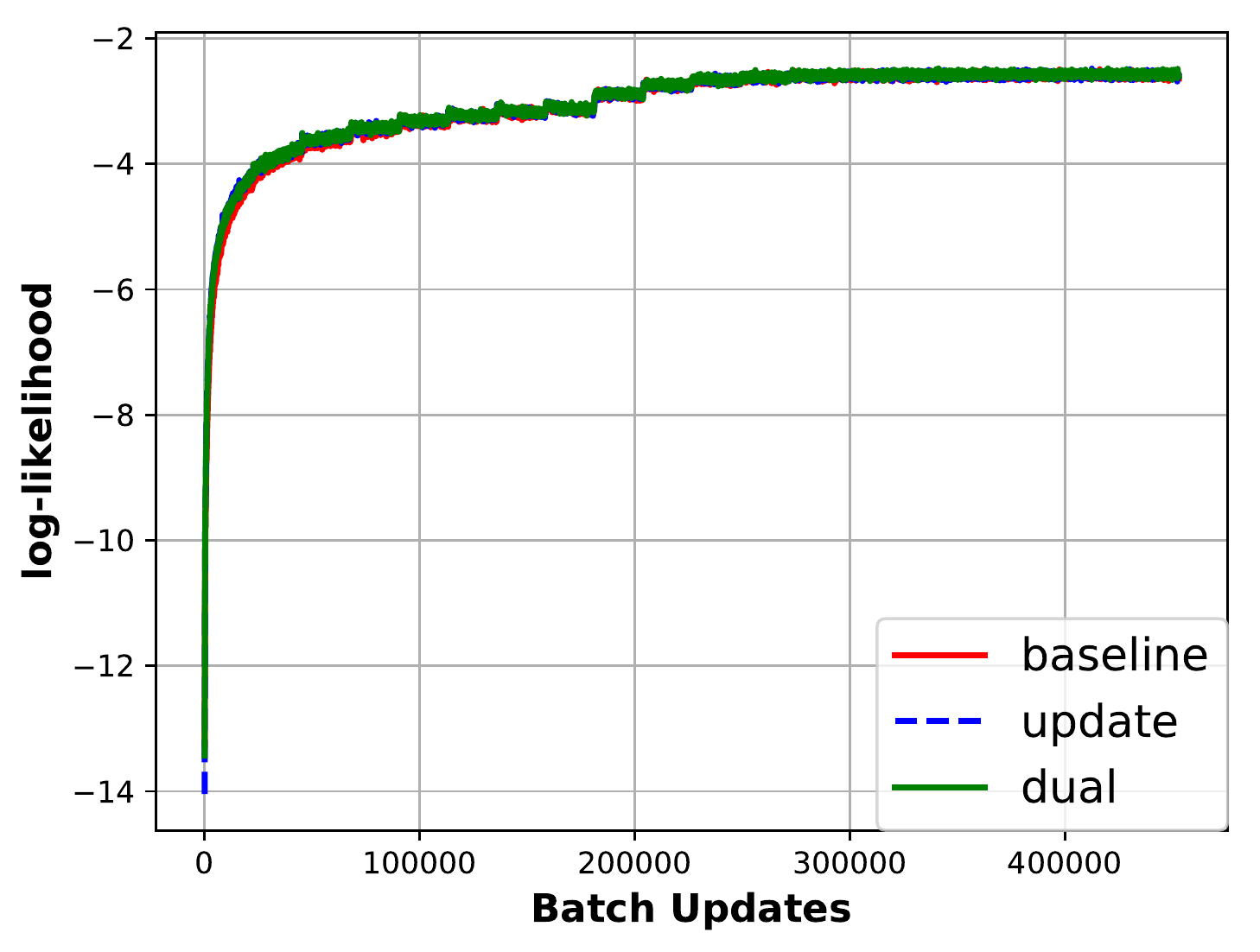}
\caption{Training Curve on Unsampled Data}
\end{subfigure}
\begin{subfigure}[b]{0.40\textwidth}
\centering
\includegraphics[width=\textwidth]{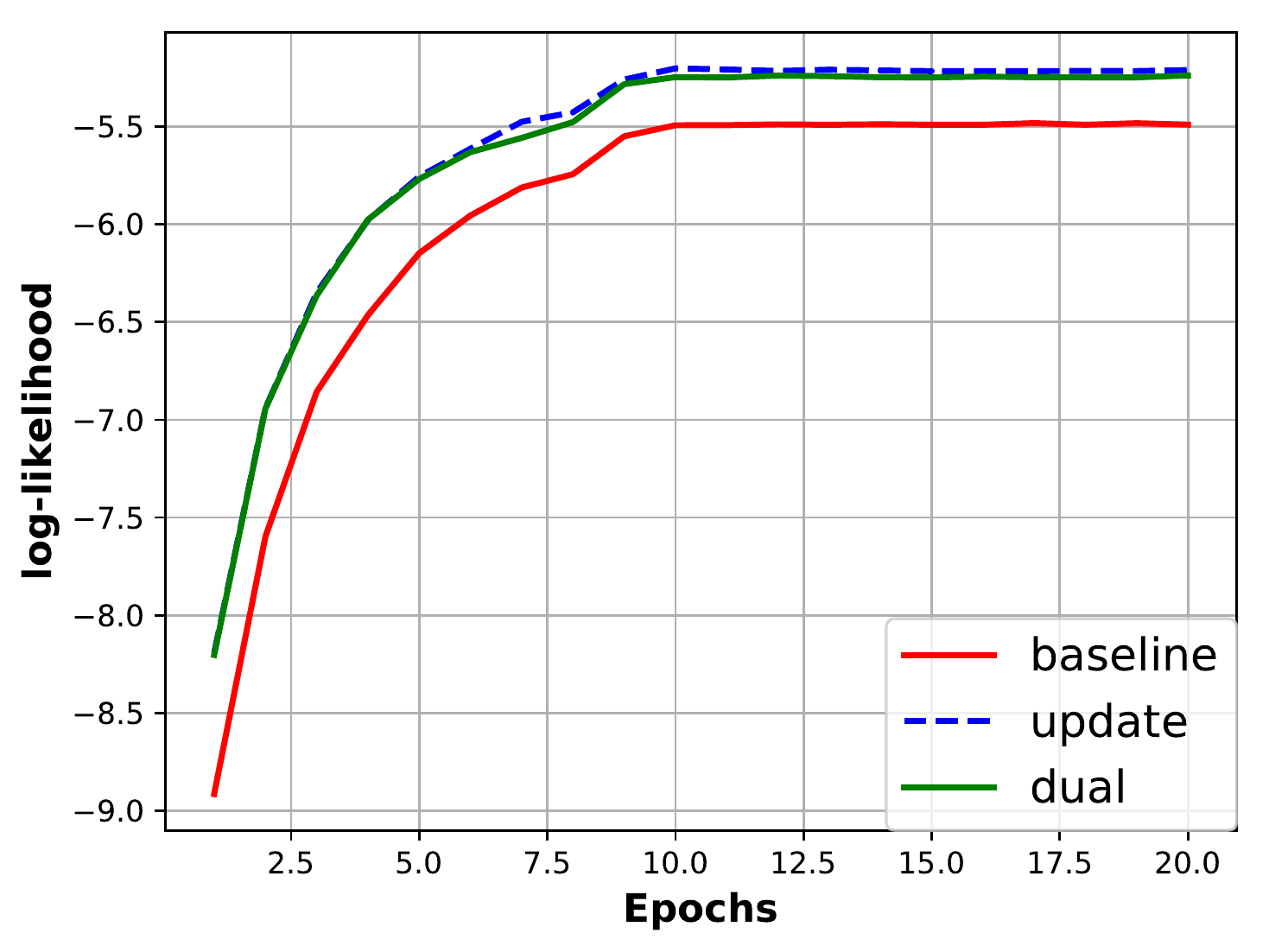}
\caption{Development Curve on 100k Sample}
\end{subfigure}
\begin{subfigure}[b]{0.40\textwidth}
\centering
\includegraphics[width=\textwidth]{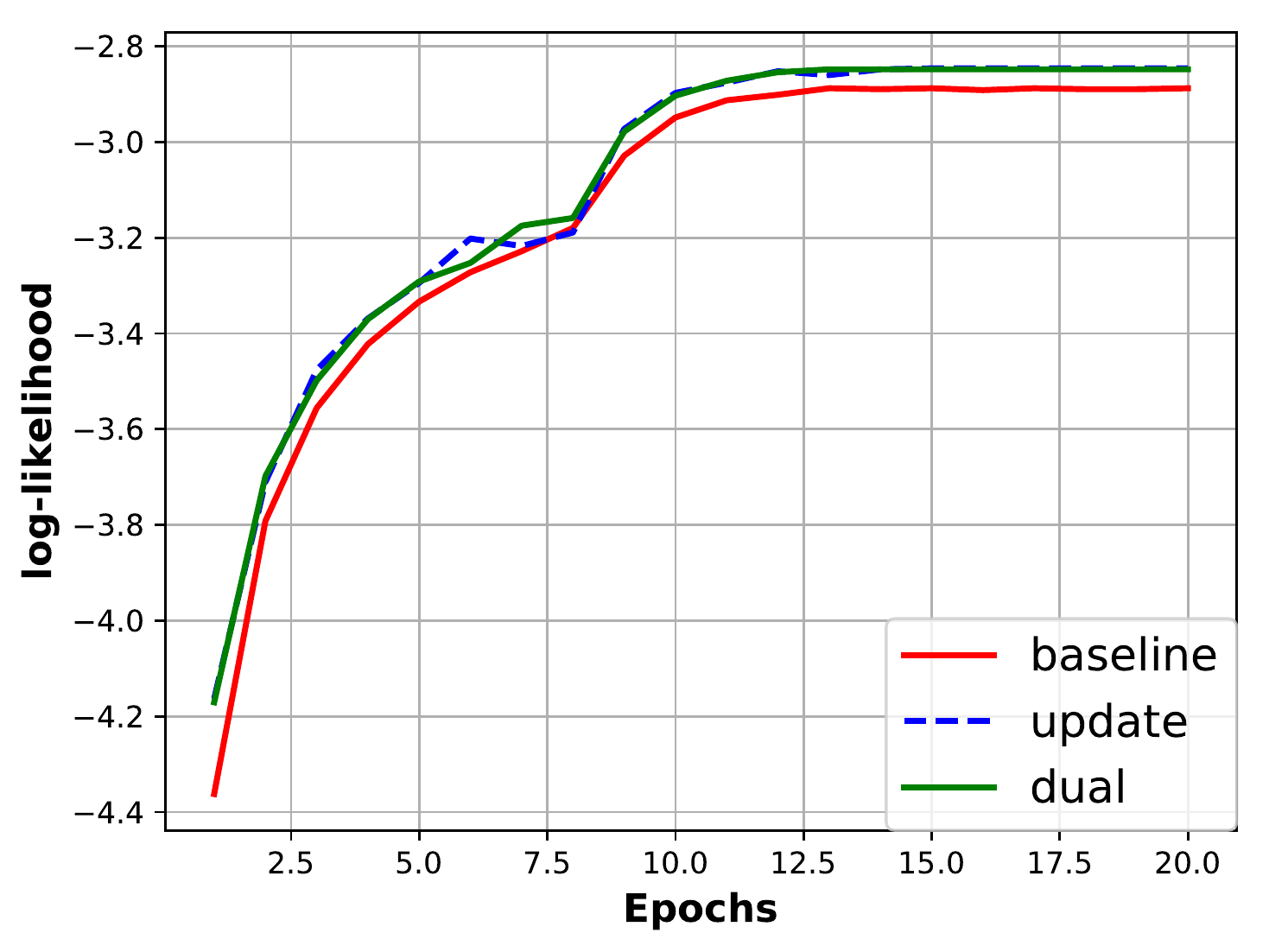}
\caption{Development Curve on Unsampled Data}
\end{subfigure}
\caption{Plot of objective function vs batch updates / epochs for Chinese-English training data
\label{fig:2}}
\end{figure*}

\begin{table*}[ht]
\centering
\begin{tabular}{c|c|ccc|ccc}
\multirow{2}{*}{Sentence Pairs} & \multirow{2}{*}{Baseline} & \multicolumn{3}{c|}{small embedding}         & \multicolumn{3}{c}{extended embedding}      \\ \cline{3-8} 
                                &                           & fixed & update           & dual           & fixed & update           & dual           \\ \hline
100,000                         & 15.99                     & 15.36$\dagger$      & 16.44          & \textbf{16.89}$\dagger$ & 15.90      & 17.55$\dagger$          & \textbf{17.56}$\dagger$ \\ \hline
250,000                         & 21.72                     & 22.05      & \textbf{22.16} & 22.08          & 22.05      & 23.04$\dagger$          & \textbf{23.38}$\dagger$ \\ \hline
500,000                         & 26.28                     & 25.48$\dagger$      & \textbf{26.51} & 26.29          & 26.13      & \textbf{27.09}$\dagger$ & 26.76$\dagger$          \\ \hline
1,000,000                       & 29.75                     & 29.54      & 29.59          & \textbf{29.83} & 29.39      & 30.6$\dagger$           & \textbf{30.89}$\dagger$ \\ \hline
2,013,142                       & 32.44                     & 31.40$\dagger$      & \textbf{33.21}$\dagger$ & \textbf{33.21} & 32.61      & 33.26$\dagger$          & \textbf{34.03}*$\dagger$ \\
\end{tabular}
\caption{Chinese-English experiment Results with Different Data Sizes. Asterisks are appended when the difference between update and dual method translation output is significant, and daggers are appended when difference between any system output and the baseline is significant. The significance level is $p < 0.05$.}
\label{tab:2}
\end{table*}

\begin{table*}[ht]
\centering
\begin{tabular}{c|c|ccc|ccc}
\multirow{2}{*}{Sentence Pairs} & \multirow{2}{*}{Baseline} & \multicolumn{3}{c|}{small embedding}         & \multicolumn{3}{c}{extended embedding} \\ \cline{3-8} 
                                &                           & fixed & update           & dual           & fixed   & update   & dual            \\ \hline
100,000                         & 13.55                     & 14.02$\dagger$      & 14.75$\dagger$          & \textbf{14.91}$\dagger$ & 15.14$\dagger$        & 15.28$\dagger$  & \textbf{15.96}*$\dagger$  \\ \hline
250,000                         & 20.11                     & 20.18      & \textbf{20.92}*$\dagger$ & 20.47          & 20.74$\dagger$        & 20.76$\dagger$  & \textbf{20.95}$\dagger$  \\ \hline
500,000                         & 23.09                     & 23.46      & \textbf{24.14}$\dagger$ & 24.11$\dagger$          & 23.55$\dagger$        & 24.14$\dagger$  & \textbf{24.64}*$\dagger$  \\ \hline
1,000,000                       & 26.17                     & 25.95      & \textbf{26.29} & 26.21          & 26.27        & 26.19  & \textbf{26.59}*$\dagger$  \\ \hline
4,562,102                       & 29.71                     & 28.73$\dagger$      & 30.00$\dagger$          & \textbf{30.07} & 29.17$\dagger$        & 29.13$\dagger$  & \textbf{30.00}*  \\
\end{tabular}
\caption{German-English experiment Results with Different Data Sizes. Asterisks are appended when the difference between update and dual method translation output is significant, and daggers are appended when difference between any system output and the baseline is significant. The significance level is $p < 0.05$.}
\label{tab:3}
\end{table*}

\subsection{Results}

Figure \ref{fig:2} shows the training and development log-probability curve under 100k and unsampled Chinese-English training data\footnote{The same plots were generated for German-English experiments and small embedding incorporations as well, and the trend looks very similar. Hence they are omitted to avoid repetitiveness.}, respectively. We can see from the figure that adding pre-trained word embedding does not speedup convergence of the training process, which verifies the conclusion from \cite{abdou2017variable}. However, our result extends over the previous work to show that while there is no change in the convergence speed, the objective function value (especially the development loss function value) at convergence is significantly higher when pre-trained word embedding is incorporated, which indicates stronger model is learned. We also notice that contrary to the general belief, pre-trained embedding initialization does not make the initial objective value significantly higher. This implies that fixing word embedding throughout the training process as what has been done in some other NLP literatures may not be a very good strategy for neural machine translation. The hypothesis is further verified by the result that follows.

% fixed vs. updated
Table \ref{tab:2} and Table \ref{tab:3} show the uncased BLEU scores and the significance of difference between some pairs of results. Comparing the results of fixed strategy and others, it can first be noticed that fixed initialization does not only almost consistently generate worst BLEU score among the three, but also significantly hurts the performance over the baseline sometimes. This strengthens our hypothesis above. On the other hand, if we allow the source-side word embedding to be updated, we can observe improvements over the baseline more often. We hence conclude that pre-trained source-side monolingual word embedding cannot directly benefit NMT performance, and adjustment by NMT training is necessary for it to be beneficial to the NMT system performance.

Comparing the Chinese-English and German-English experiments, we notice that incorporating embeddings for German-English experiments yields significant improvements over baselines more often, mainly because of the improvements obtained by extended fixed embedding incorporation. We think this is due to fact that the German-English test set generally has higher pre-BPE OOV rate across different sample size of the parallel training data.

% small vs. full
Comparing the results under different monolingual data scales, for Chinese-English experiments, the extended embedding always performs better than small embedding, while for German-English experiments, we got mixed results when comparing results with two different kinds of embeddings. This can be explained by the fact that the parallel training data has a minor domain mismatch from the monolingual training data (parliament proceedings vs. news). In terms of incorporation strategy, it can be observed that while the dual strategy is not very helpful with small word embeddings, improvements over update strategy can almost always be obtained with extended embeddings. We can also notice that under the German-English setting with extended embeddings, the different between update and dual embedding method is more often significant compared to that of Chinese-English experiments, signaling that dual embedding method is more robust to domain variances between monolingual data and bi-text.

% different amount of data
% dual vs. plain
Comparing the results obtained under different parallel training data scales, it can be observed that the benefit of source-side monolingual word embedding (compared to the baseline) seems to be decreasing along with the increasing amount of data, verifying the intuition that extra monolingual information is most useful under low-resource settings. On the other hand, the dual embedding model is able to obtain the largest performance gains over update initialization both for Chinese-English and the German-English training data with extended embeddings. This indicates that the dual embedding model is able to get the best of both worlds as expected -- it leverages more on the initialized word embedding at low resource settings, but is able to learn useful supplementary features when relatively large amount of parallel training data is available.

\subsection{Analysis}

% \subsubsection*{Qualitative Analysis}

\begin{table*}[ht]
\centering
\begin{tabular}{|c|}
\hline
% \begin{tabular}{p{\linewidth}}{@{}l@{}}Src: \begin{CJK*}{UTF8}{gbsn} 汪@@ 世@@ 林 说 , 泛@@ 美 体育 组织 各 成员国 家 和 地区 都 对 参加 北京 奥运会 表现 出 极大 的 热情 , 全部 42 个 成员 都 已 确认 将 参加 下 个 月 在 北京 召开 的 奥运会 各 国 和 地区 奥@@ 委会 \end{CJK*} \\ \begin{CJK*}{UTF8}{gbsn} 代表团 团长 会议 . \end{CJK*} \\ Update: All 42 members of the member states of the United States and the United States have confirmed their participation in the Beijing Olympic Games in Beijing next month , Wang said .\\ Dual: Wang said that all the members and regions of the Pan @-@ American Sports Organization have shown great enthusiasm for participating in the Beijing Olympic Games , and all 42 members have confirmed that they will attend the meeting of various countries and regions in Beijing next month .\\ ref: Wang Shilin said that all member nations and regions of the Pan @-@ American Sports Organization had displayed great enthusiasm for the participation of the Beijing Olympics . All 42 members had confirmed their participation in the meeting of the Chef de Mission of the National Olympic Committees to be held in Beijing next month .\end{tabular} \\
% \hline
{\small
\begin{tabular}{p{0.01\linewidth}p{0.90\linewidth}}
% \begin{tabular}
1 & \textbf{Source 1}: \begin{CJK*}{UTF8}{gbsn} Len@@ ovo\ 即将\ 要\ 在\ 7月\ 17日\ (\ 美国\ 时间\ )\ 推出\ 的\ \colorbox{pms1215c}{Think@@\ Pad\ T@@\ 61\ p}\ ,\ 也是\ \colorbox{pms584c}{T\ 系列}\ 最@@\ 高级\ 的\ 笔记本\ 电脑\ ,\ 给\ 我们\ 一\ 个\ 相当\ 不错\ 的\ 惊喜\ :\ \colorbox{pms564c}{U@@\ W@@\ B\ (\ Wis@@\ air\ } \colorbox{pms564c}{Ul@@\ tra\ Wi@@\ deb@@\ and\ )}\ .\ \end{CJK*} \\
2 & \textbf{Reference 1}: The \colorbox{pms1215c}{ThinkPad T61p} that Lenovo is about to introduce on July 17 ( U.S. time ) , also the most advanced notebook computer of the \colorbox{pms584c}{T series} , has given us a very pleasant surprise : \colorbox{pms564c}{UWB ( Wisair Ultra Wideband )} . \\
3 & \textbf{Baseline 100k}: On July 17 ( US time ) launched , the number of SWB's highest notebook notebook computers , which is also a good source of notebook notebook computers . : \colorbox{pms564c}{UWB} .\\
4 & \textbf{Update 100k}: Stewart is scheduled to be launched on July 17 ( the US time ) to launch a very good surprise of the top - - - - - - - - - - - - - - - - - - - - - - - - - - - - ranking computer computer to us .\\
5 & \textbf{Dual 100k}: Founded to July 17 , Inc will be launched to be launched on the current model notebook , which is also a well - - - - - - - - - - - - - - - - - - - - - - - - - - - - - - - - - - - - - - - - - - - - - - - - - - - - - - - - - - - - - - - - - - - - - - - - - \\
6 & \textbf{Baseline Unsampled}: It will be launched on July 17 ( USA time ) as a top-class notebook computer and gives us a rather good surprise: \colorbox{pms564c}{UWB Ultra Witra Wireless}. \\
7 & \textbf{Update Unsampled}: It is about to be launched on July 17 ( US time ) , as well as the \colorbox{pms584c}{T}'s most advanced notebooks of notebooks, to us: \colorbox{pms564c}{UWB (Wisair Ultra Wireless and)} .\\
8 & \textbf{Dual Unsampled}: The Dell \colorbox{pms1215c}{Pad T6p} , which is about to be launched on July 17 ( US time ) , is also the \colorbox{pms584c}{T-series}'s most advanced notebooks , giving us a rather good surprise: \colorbox{pms564c}{UWB ( Wisair Ultra Wideband )} .\\
\hline\hline
9 & \textbf{Source 2}: \begin{CJK*}{UTF8}{gbsn} 由\ \colorbox{pms1215c}{王@@\ 兵@@\ 兵 黑\ 砖@@\ 窑\ 案}\ 引发\ 的\ \colorbox{pms584c}{山西\ 黑\ 砖@@\ 窑\ 奴@@\ 工\ 事件}\ ,\ 曾\ 国内外\ 一度\ 引起\ 关注\ ,\ 中央\ 高层\ 批示\ 要求\ 严查\ .\ \end{CJK*}\\
10 & \textbf{Reference 2}: The \colorbox{pms584c}{Shanxi black brick kiln slave labor incident} touched off by the \colorbox{pms1215c}{black brick kiln case of} \colorbox{pms1215c}{Wang Bingbing} once attracted attention from inside the country and abroad . The top leadership of the central government had given directive demanding stern prosecution . \\
11 & \textbf{Baseline 100k}: From \colorbox{pms1215c}{the incident caused by Wang Wei}, \colorbox{pms584c}{the Shanxi case caused by Wang Wei}, and from home and abroad , and his attention to the high - level instructions of the central authorities .\\
12 & \textbf{Update 100k}: In \colorbox{pms1215c}{the case of Wang Shanxi} , \colorbox{pms584c}{the Shaanxi Shuan case of Shanxi Province in Shanxi Province }, has attracted great attention to the party and abroad . .\\
13 & \textbf{Dual 100k}: In \colorbox{pms1215c}{the case of the Shanxi} , the government of \colorbox{pms584c}{Shanxi's Shanxi Shengsheng incident} , has once again attracted attention from home and abroad .\\
14 & \textbf{Baseline Unsampled}: At a time , the central authorities \&apos; instructions have aroused concern at home and abroad , and the central authorities have issued instructions to investigate the incident . \\
15 & \textbf{Update Unsampled}: At one point at home and abroad , there was a great deal of attention at home and abroad , and the central authorities demanded a strict investigation .\\
16 & \textbf{Dual Unsampled}: The \colorbox{pms584c}{incident} , which was triggered by the \colorbox{pms1215c}{black brick kiln of Wang Wei - bing} , has once aroused concern at home and abroad , and the central authorities ' high - level instructions have been set for investigation .
\end{tabular}} \\
% \hline
% begin{tabular}{p{\linewidth}}{@{}l@{}}Src: \begin{CJK*}{UTF8}{gbsn} 随着 旅游 旺季 的 到来 , 给 天@@ 南@@ 地@@ 北 的 游客 创造 一 个 文明 的 旅游 环境 , 应该 说 是 当务之急 . \end{CJK*} \\ Baseline: With the arrival of the tourism season , creating a civilized tourism environment for tourists from the north and south should be said to be an urgent task . \\ Update: With the arrival of the tourist season , creating a civilized tourism environment for tourists in northern Tibet should be said to be an urgent task .\\ Dual: With the arrival of the peak travel season , it should be said that the most urgent task is to create a civilized tourism environment for tourists from the north and south .\\ ref: With the coming of the peak travel season , it should be said that creating a civilized travel environment for tourists from near and far is an urgent matter .\end{tabular} \\
\hline
\end{tabular}
\caption{Two translation snippets from the 100k sample and unsampled Chinese-English experiments with several named entities highlighted in corresponding colors. All the embedding incorporated are extended embeddings. The @@ symbol is the token breaking symbol produced by BPE processing.}
\label{tab:4}
\end{table*}

\subsubsection*{Qualitative Analysis}

% \begin{wraptable}{r}{5cm}
\begin{table}
\centering

\begin{tabular}{l|l}
system   & accuracy \\ \hline
baseline & 29.10\%  \\ \hline
update   & 32.09\%  \\ \hline
dual     & \textbf{33.58\%} 
\end{tabular}
\caption{Human-evaluated singleton word translation accuracies on Chinese-English test set.\label{tab:5}}
\vspace{-0.3cm}
\end{table}
% \end{wraptable}

Translation of test sets were examined manually to evaluate the qualitative improvements obtained by incorporating pre-trained word embeddings within NMT. In the case of small embedding, we have learned from significance test that many of these improvements are not statistically significant. But even for the significant ones, we didn't observe very specific patterns for qualitative improvements.

In the case of extended embedding, however, we observed an specific improvement of the translation adequacy for rare words (mainly named entities) in the training data, and the usage of dual embedding model often brings further improvements in that aspect. Such improvement is evident in Table \ref{tab:4} when comparing the translation results from systems trained with unsampled parallel data (sentence 6-8 and sentence 14-16).

To further verify the observation above, we did a simple human evaluation where we take the translation output from Chinese-English systems trained on unsampled parallel data with extended embedding incorporation. We first took the singleton words (before BPE) in the unsampled parallel data and filtered the test sentences containing these singleton words. We then manually read the filtered and shuffled test sentences and answered the yes/no question: {\it are the singleton words appeared in the sentence translated correctly in the test output}? We chose to analyze singleton words rather than OOV words because (1) the translation of singleton words is less noisy; (2) if a word occurs in both the training set and the test set, it's more likely to occur in the monolingual data and hence embedding incorporation will add extra information to translate these words. We found 226k singleton words in the training data and 134 occurrences of these words in the test data. The results is shown in the Table \ref{tab:5} and we can see that both update and dual incorporation method improves the singleton translation accuracy. This agrees with our observation as presented in Table \ref{tab:4}.

On the other hand, while the BLEU scores improves significantly over baselines under the low-resource settings with extended embedding incorporation, we did notice that the systems with embedding incorporated tends to produce repetitive output more frequently (e.g. sentence 4 and 5), to which we don't have a very good explanation. We conjecture that such problem could be remedied by coverage modeling techniques such as \cite{Tu:2016ut} and \cite{wu2016google}, but leave the verification of it as future work. We also acknowledge that the improvements on rare word translation is not obvious under low-resource settings (e.g. sentence 11-13) because translation outputs are often too noisy to tell much useful qualitative trend.

% \subsubsection*{Quantitative Analysis}

% TODO: correct the numbers for the updates.

\begin{figure*}
\begin{subfigure}[b]{0.49\textwidth}
\centering
\includegraphics[width=\textwidth]{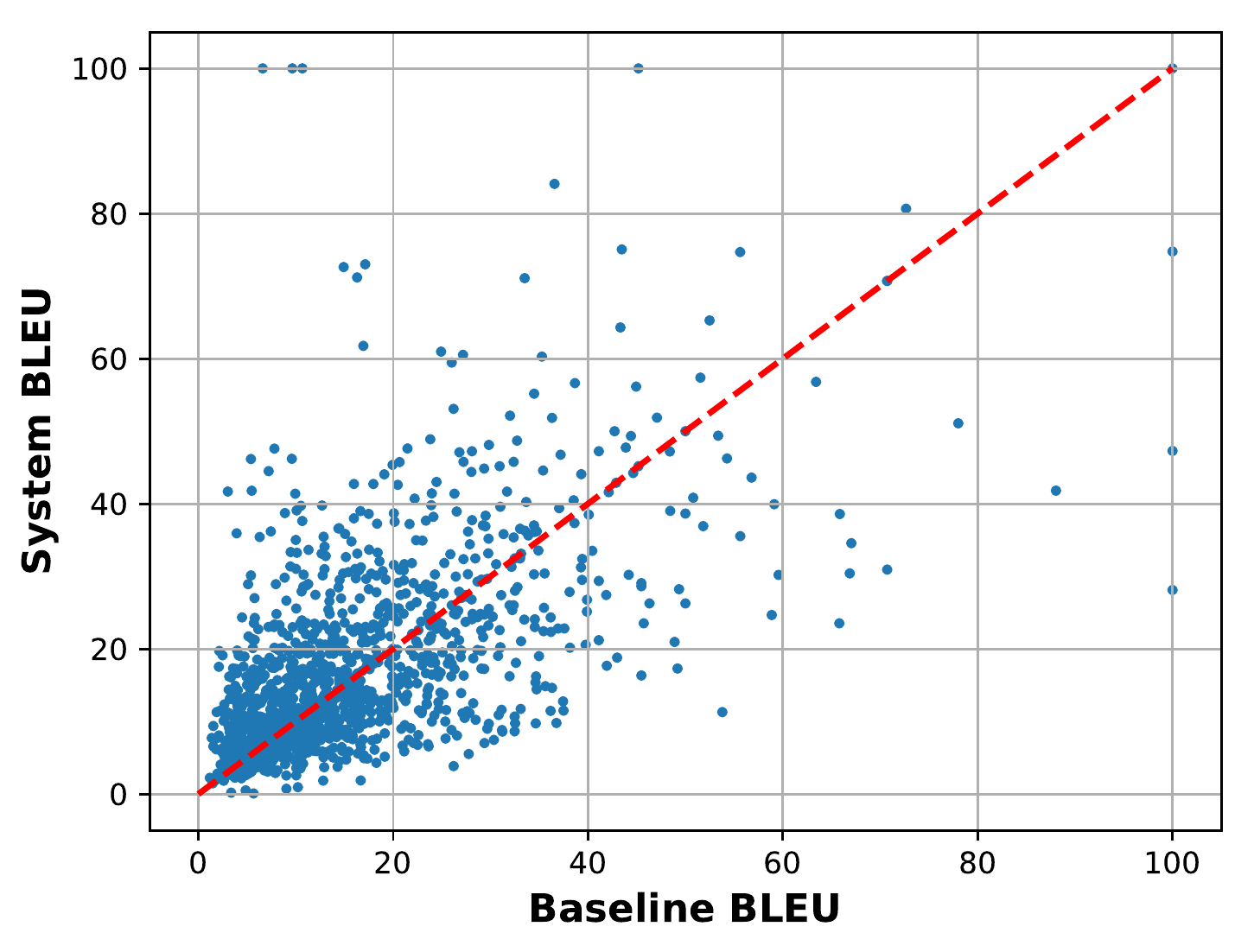}
\caption{ZH-EN, Extended Update Embedding}
\end{subfigure}
\begin{subfigure}[b]{0.49\textwidth}
\centering
\includegraphics[width=\textwidth]{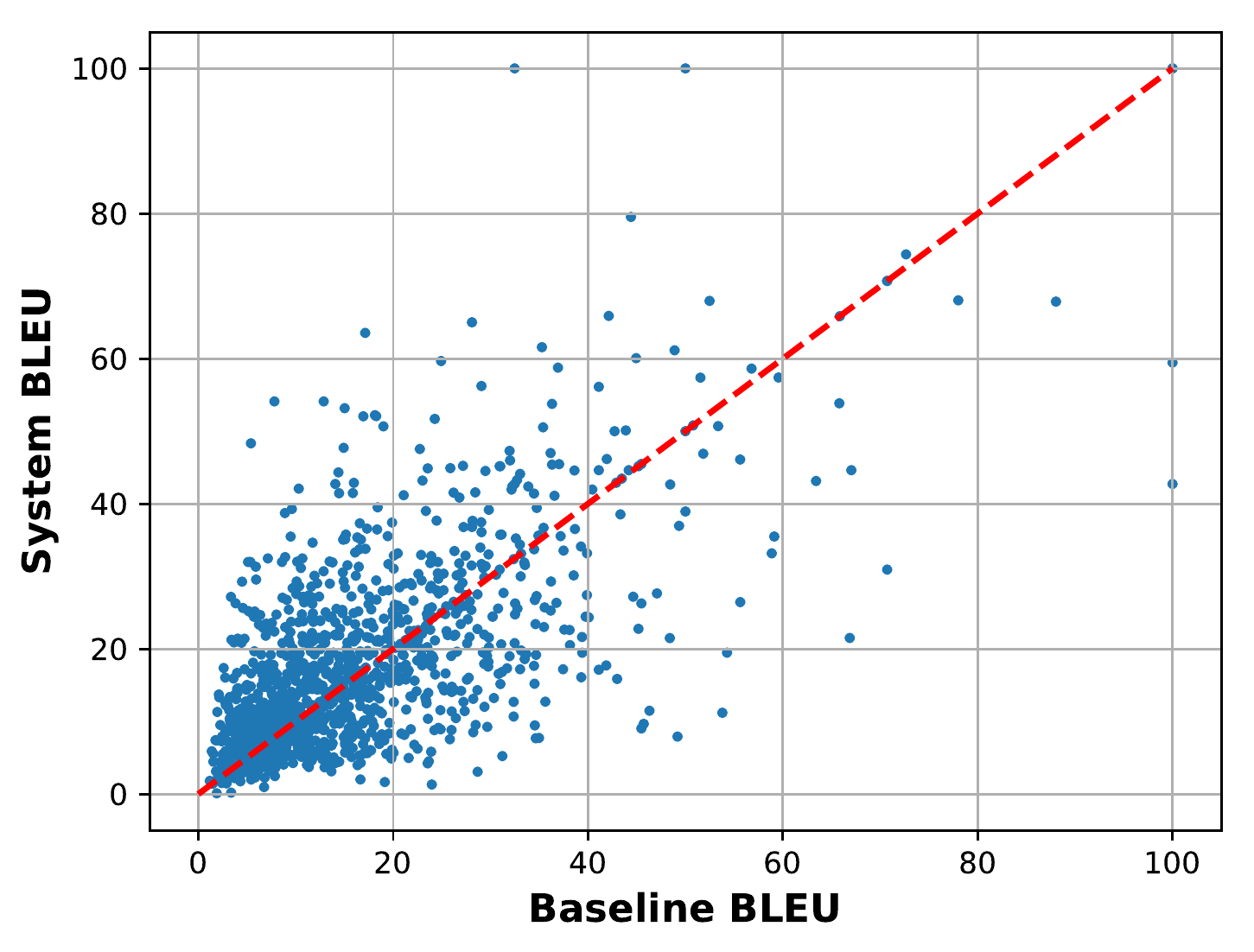}
\caption{ZH-EN, Extended Dual Embedding}
\end{subfigure}
\begin{subfigure}[b]{0.49\textwidth}
\centering
\includegraphics[width=\textwidth]{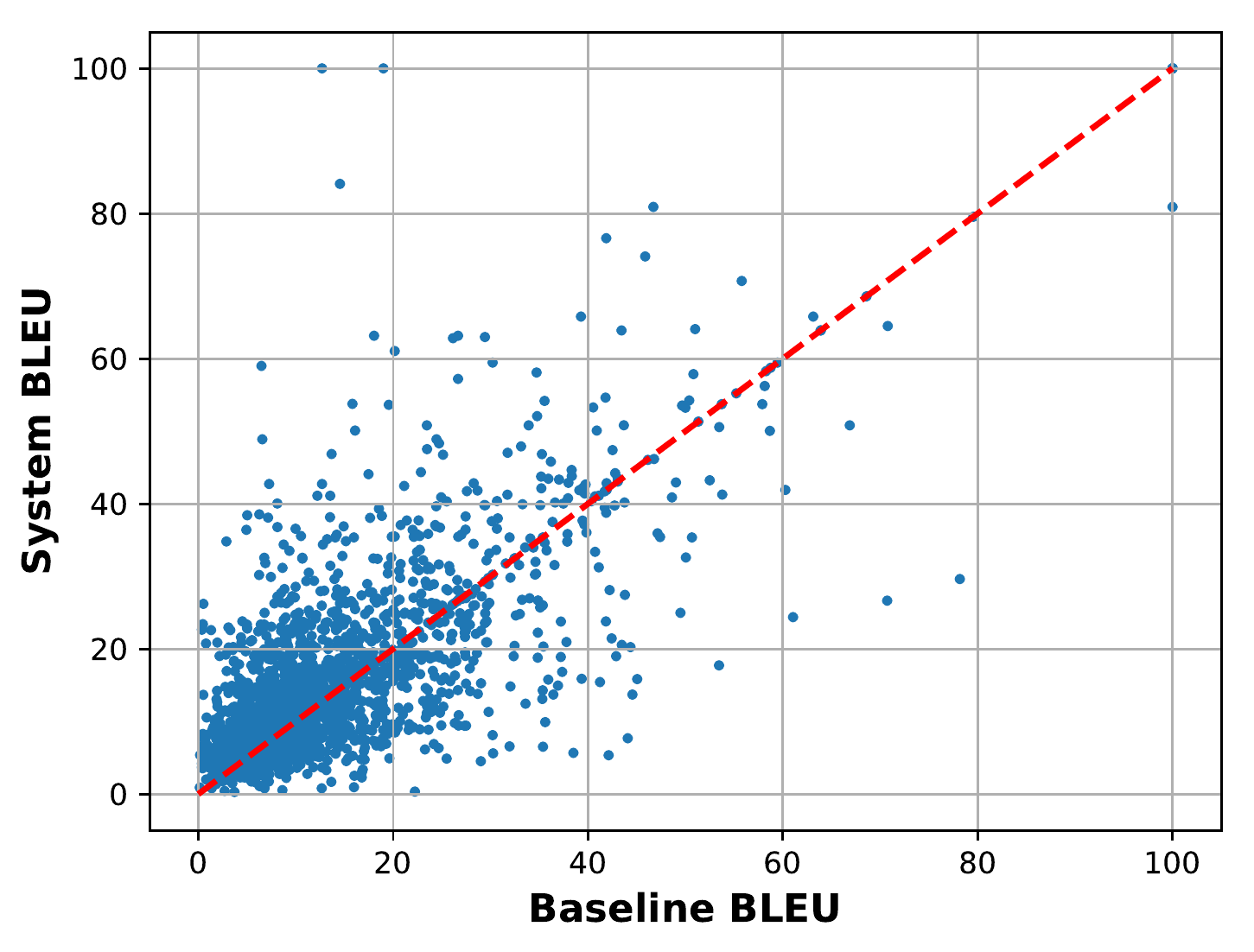}
\caption{DE-EN, Extended Update Embedding}
\end{subfigure}
\begin{subfigure}[b]{0.49\textwidth}
\centering
\includegraphics[width=\textwidth]{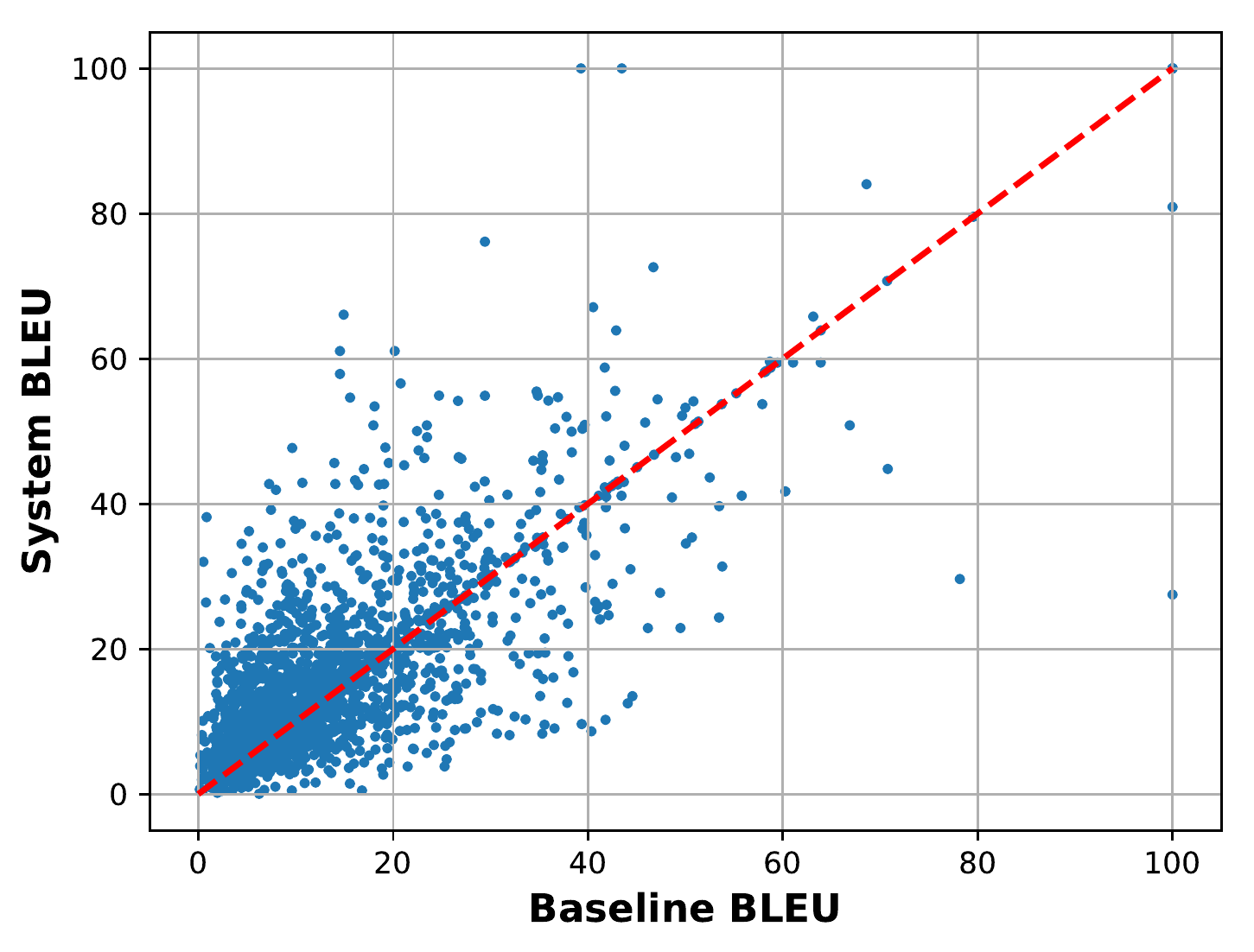}
\caption{DE-EN, Extended Dual Embedding}
\end{subfigure}
\caption{Sentence-Level BLEU Scatter Plot Between Baseline and Embedding-Incorporated Systems. The dots on the upper-left part of the red line corresponds to system output sentences that are better than the baseline, and vice versa.
\label{fig:3}}
\end{figure*}

\begin{figure*}
\begin{subfigure}[b]{0.49\textwidth}
\centering
\includegraphics[width=\textwidth]{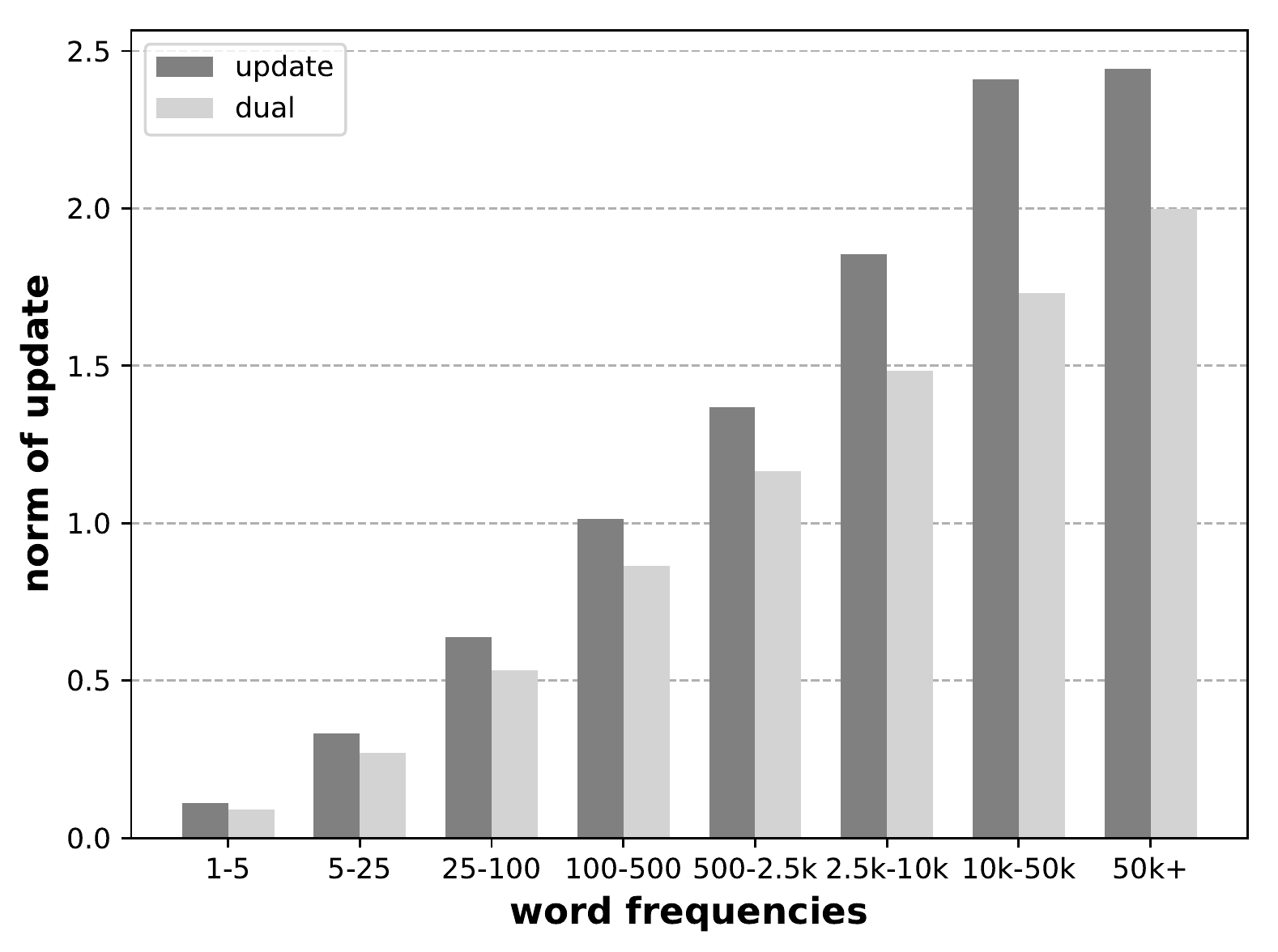}
\caption{ZH-EN}
\end{subfigure}
\begin{subfigure}[b]{0.49\textwidth}
\centering
\includegraphics[width=\textwidth]{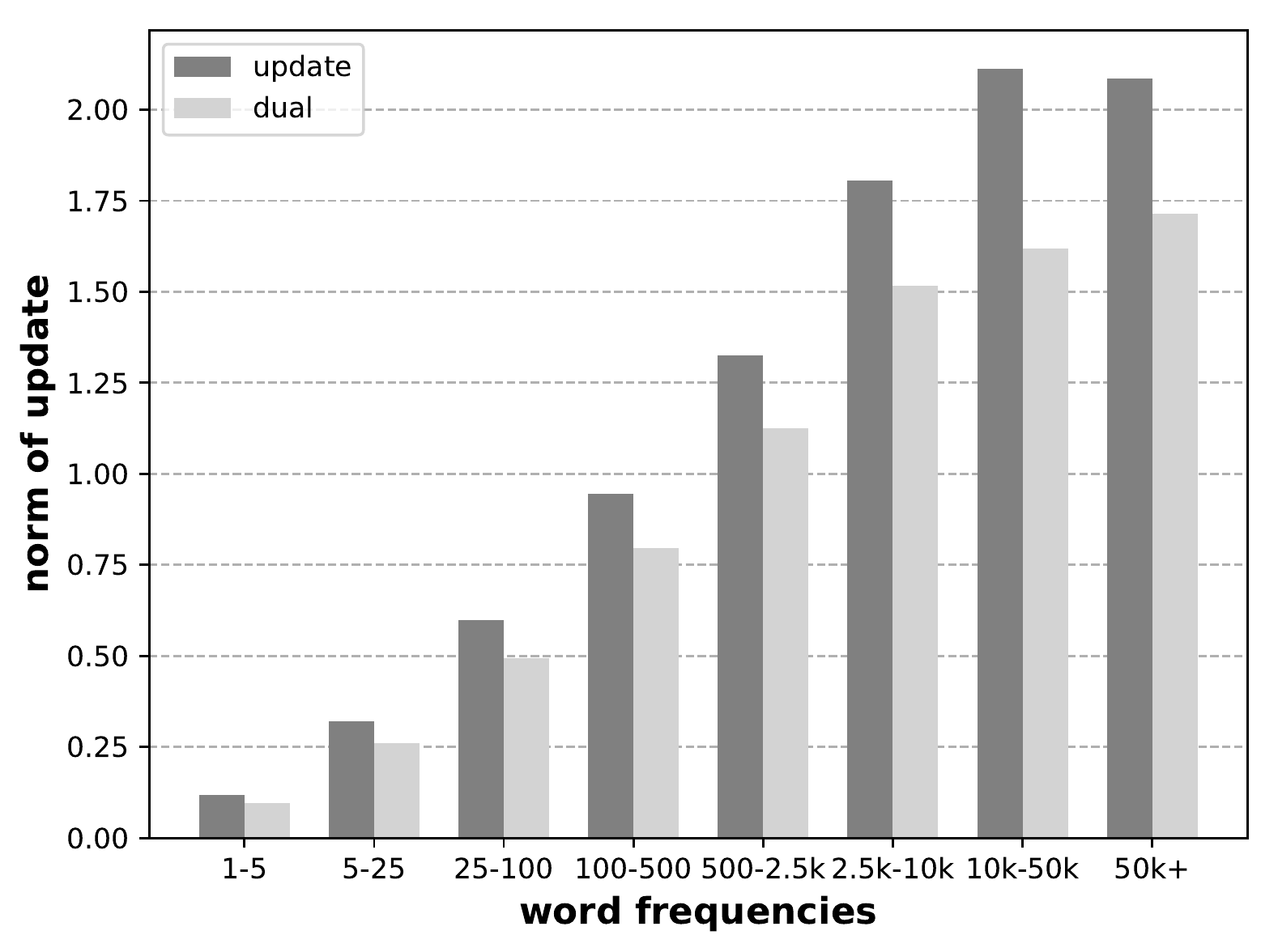}
\caption{DE-EN}
\end{subfigure}
\caption{Average Norm of Update on Word Embeddings Grouped by Word Frequency in Training Data
\label{fig:4}}
\end{figure*}

\subsubsection*{Quantitative Analysis}

For quantitative analysis, we focus on the small-scale experiment (trained with 100k sample of parallel data) with extended embedding incorporation as they seem to pose most interesting improvements in terms of BLEU scores.

We started with computing the sentence-level BLEU with MultEval toolkit \cite{clark2011better} and generating scatter plot of sentence-level BLEU score of the update and dual embedding system against the baseline system for each output sentence decoded on the test set, as shown in Figure \ref{fig:3}. The purpose of this analysis is to examine the variance of output sentence before/after the embedding incorporation. It could first be noticed that across all embedding incorporation methods, the dots are shifted to the upper left side of the red line, which means sentence-level BLEU score tends to increase after incorporation. This agrees with the increase of corpus-level BLEU score in Table \ref{tab:2} and Table \ref{tab:3}. On the other hand, all embedding incorporation methods incurs similar amount of variances in translation outputs, even on different language pairs. In terms of comparison across incorporation strategies, the update strategy seems to incur slightly more drastic BLEU scores changes (dots that are close to the right and upper part of the horizontal and vertical axes, respectively), but the difference is not significant enough to make a strong argument.

Another problem we are interested in is the norm of update on the word embeddings during the NMT training process. More specifically, for each words in the dictionary, we take their word embedding before/after the training process and compare the norm of their difference. Figure \ref{fig:4} shows the norm of update grouped by the word frequency in the training data. It should be noted that the norm of update is increasing roughly linearly along with the frequency of words up till 50000. This implies that for each iteration, unless the word has been seen extremely frequently, the norm of update performed on the word embedding is about the same on average. We also see that norm of update performed on dual incorporation strategy is consistently lower than update incorporation strategy. Because the pre-trained part of the embedding is fixed, and the dual strategy is essentially learning a correction term over the pre-trained word embedding rather than rewriting the pre-trained value completely, we conjecture that the fixed part of the dual embedding is preventing the updated part of the dual embedding to perform too much correction over its pre-trained value. This conservativeness in performing update may account for the extra robustness of dual embedding incorporation we observed in the qualitative analysis.

\section{Conclusion} \label{sec:conclusion}

Our analysis on using source-side monolingual word embeddings in NMT indicates that (1) the source-side embeddings should be updated during NMT training; (2) the source-side embeddings are more effective when bilingual training data is limited, especially when OOV rates is high. Moreover, source-side embedding incorporation is also useful under some high-resource settings when incorporated properly; (3) the effect of source-side word embedding strengthens when extra monolingual data is provided for training, and the domain of the monolingual data also seems to matter.

We recommend that incorporating pre-trained embeddings as input become a standard practice for NMT when bilingual training data is scarce, especially when extra source-side monolingual data is available. While incorporating pre-trained embeddings at high-resource settings may also be helpful, we advise that extra caution should be used to ensure the monolingual data is in-domain, and appropriate incorporation strategy should be selected.

% \bibliography{all}
% \bibliographystyle{acl_natbib} % naacl
% \bibliographystyle{acl} % coling

\bibliographystyle{acl_natbib} % acl

\end{document}